\documentclass[runningheads]{llncs}
\usepackage[T1]{fontenc}
\usepackage{booktabs}
\usepackage{graphicx} % for \resizebox
\usepackage{hyperref}

\usepackage{amsmath}
\usepackage{amssymb}
\usepackage{multirow}
\usepackage{multicol}
\usepackage{marvosym}

\usepackage[table]{xcolor} 
\definecolor{tabHeader2}{HTML}{EDEDED}   % very 
\definecolor{ourbg}{HTML}{E0F2FE}   
\definecolor{gbg}{HTML}{DCFCE7}

\definecolor{ours}{RGB}{235,245,255}      % light blue
\definecolor{groupgray}{gray}{0.96}       % very light gray

\usepackage{graphicx,verbatim}
\newsavebox{\biogaittitleiconbox}
\sbox{\biogaittitleiconbox}{%
  \includegraphics[height=1.7cm]{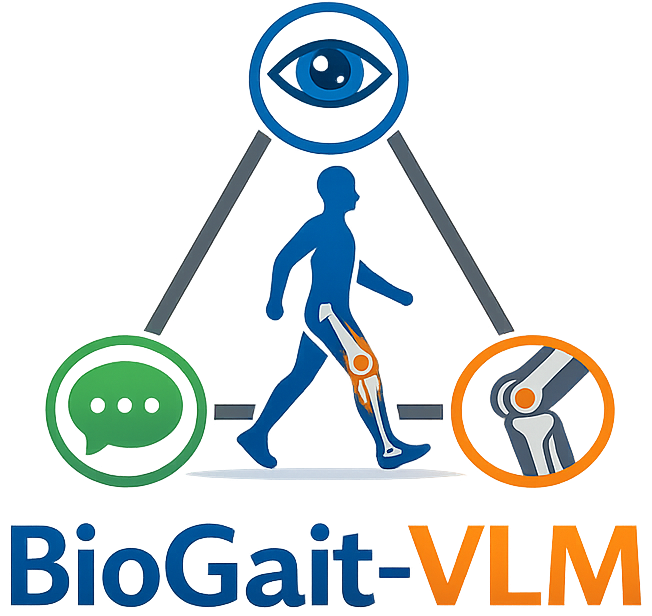}%
}
\DeclareRobustCommand{\BioGaitTitleIcon}{%
  \raisebox{\dimexpr-3\ht\biogaittitleiconbox/4\relax}{%
    \usebox{\biogaittitleiconbox}%
  }%
}

\makeatletter
\renewcommand{\paragraph}{%
  \@startsection{paragraph}{4}%
  {\z@}{0.2ex \@plus 1ex \@minus .2ex}{-1em}%
  {\normalfont\normalsize\bfseries}%
}
\makeatother

\begin{document}
%
%\title{BioGait-VLM: Vision–Language–Biomechanics Fusion for Clinical Gait Understanding}

%\title{BioGait-VLM: Biomechanically Grounded Vision-Language Models for Interpretable Clinical Gait Assessment}

% \title{BioGait-VLM: A Tri-Modal Vision-Language-Biomechanics Framework for Interpretable Clinical Gait  Assessment}

% \title{\texorpdfstring{%
% \begin{tabular}{@{}c@{}p{0.90\textwidth}@{}}
% \raisebox{-0.75\height}{\includegraphics[height=1.7cm]{fig/biogait_icon.png}} \hspace{-1.1em}  &
% \centering A Tri-Modal Vision--Language--Biomechanics\\
% Framework for\\
% Interpretable Clinical Gait Assessment
% \end{tabular}
% }{BioGait-VLM: A Tri-Modal Vision--Language--Biomechanics Framework for Interpretable Clinical Gait Assessment}}

\title{\texorpdfstring{%
\begin{tabular}{@{}c@{}p{0.90\textwidth}@{}}
\BioGaitTitleIcon \hspace{-1.1em}  &
\centering A Tri-Modal Vision--Language--Biomechanics\\
Framework for\\
Interpretable Clinical Gait Assessment
\end{tabular}
}{BioGait-VLM: A Tri-Modal Vision--Language--Biomechanics Framework for Interpretable Clinical Gait Assessment}}

\titlerunning{BioGait-VLM}
%\begin{comment}  %% Removed for anonymized MICCAI submission
\author{Erdong Chen\inst{1,2}\thanks{ \scriptsize Equal contribution. Erdong was a visiting research intern at Drexel University during this work.}  \and
Yuyang Ji\inst{1}\textsuperscript{$\star$} \and
Jacob K. Greenberg\inst{2}  \and 
Benjamin Steel\inst{3} \and
Faraz Arkam\inst{2} \and
Abigail Lewis\inst{2} \and
Pranay Singh\inst{2} \and 
Feng Liu\inst{1}\textsuperscript{\Letter}
}
\authorrunning{E. Chen and Y. Ji et al.}
\institute{Department of Computer Science, Drexel University \and
Department of Neurological Surgery, Washington University \and University of California, Berkeley\\
}

%\end{comment}

% \author{Anonymized Authors}  %% Added for anonymized MICCAI submission
% \authorrunning{Anonymized Author et al.}
% \institute{Anonymized Affiliations \\
%     \email{email@anonymized.com}}
  
\maketitle              % typeset the header of the contribution
\begingroup
\renewcommand{\thefootnote}{} % suppress the default * / number marker
\footnotetext{\Letter\  \texttt{fl397@drexel.edu}}
\endgroup

\begin{abstract}
Video-based Clinical Gait Analysis often suffers from poor generalization as models overfit environmental biases instead of capturing pathological motion. To address this, we propose \textbf{BioGait-VLM}, a tri-modal Vision-Language-Biomechanics framework for interpretable clinical gait assessment. Unlike standard video encoders, our architecture incorporates a Temporal Evidence Distillation branch to capture rhythmic dynamics and a Biomechanical Tokenization branch that projects 3D skeleton sequences into language-aligned semantic tokens. This enables the model to explicitly reason about joint mechanics independent of visual shortcuts. To ensure rigorous benchmarking, we augment the public GAVD dataset with a high-fidelity Degenerative Cervical Myelopathy (DCM) cohort to form a unified 8-class taxonomy, establishing a strict subject-disjoint protocol to prevent data leakage. Under this setting, BioGait-VLM achieves state-of-the-art recognition accuracy. Furthermore, a blinded expert study confirms that biomechanical tokens significantly improve clinical plausibility and evidence grounding, offering a path toward transparent, privacy-enhanced gait assessment.
\href{https://vilab-group.com/project/biogait-vlm}{Project} 

\end{abstract}
%
%
%

%--------------------------------------------------------------%
\section{Introduction}

Gait is a sensitive functional biomarker of neurological and musculoskeletal health~\cite{HorakMancini2013ObjectiveBiomarkers,MonteroOdasso2017DualTaskGait,McArdle2019GaitSignatures,yin2025progait}, making clinical gait analysis (CGA) a fundamental tool for screening, severity assessment, and longitudinal monitoring of neurodegenerative diseases~\cite{zhou2024gait,li2025text,Studenski2011GaitSpeed}. Despite its diagnostic value, gold-standard CGA is restricted to controlled settings using instrumented walkways (\emph{e.g.}, GAITRite~\cite{bilney2003concurrent}) or body-worn inertial sensors. These methods require on-site setup and specialized personnel, limiting their scalability for frequent longitudinal follow-up~\cite{bilney2003concurrent,HorakMancini2013ObjectiveBiomarkers}.

To enable scalable assessment, researchers have adapted deep learning architectures to clinical gait analysis~\cite{Ranjan2025GAVD,mahfouf2025pdgv,ranjan2024developing}, such as I3D~\cite{ao2017quo} and SlowFast~\cite{Feichtenhofer2019SlowFast} to extract spatiotemporal features directly from video frames. More recently, Video Transformers like TimeSformer~\cite{bertasius2021space} and VideoMAE~\cite{tong2022videomae} have been employed to capture long-range temporal dependencies. %While these end-to-end models show promise in distinguishing broad gait categories, they inherently operate as ``black boxes,'' learning abstract feature representations that correlate with pathology but lack explicit biomechanical grounding. Consequently, they struggle to provide the \emph{granular, interpretable kinematic evidence}, such as specific joint angle deviations, that clinicians require for decision-making.
While promising, these end-to-end models operate as ``black boxes,'' learning abstract features that correlate with pathology but lack explicit biomechanical grounding. Consequently, they fail to provide the \emph{granular, interpretable kinematic evidence}, such as specific joint angle deviations, that clinicians require for decision-making.
In parallel, skeleton-based graph networks (\emph{e.g.}, ST-GCN~\cite{yan2018spatial}) have been utilized to capture geometric structure; however, these approaches typically output class probabilities without the semantic reasoning capabilities required to generate textual clinical reports.

%%
%Recently, Large Vision-Language Models (LVLMs)~\cite{zhu2025internvl3exploringadvancedtraining,liu2023llava} offer a potential solution to this interpretability gap. By aligning visual features with semantic text, models like LLaVA-Med~\cite{li2023llava} have demonstrated impressive capabilities in generating descriptive reports for static medical imaging (\emph{e.g.}, Radiology). However, adapting these general-purpose reasoning engines to \textit{dynamic} clinical gait analysis remains an open challenge. Standard LVLMs primarily rely on static visual cues and lack the domain-specific knowledge to quantify fine-grained temporal abnormalities (\emph{e.g.}, foot drop, circumduction). Furthermore, when deployed on uncontrolled ``in-the-wild'' clinical videos, these models are prone to \textit{shortcut learning}: they often attend to environmental confounders, such as a hospital bed or a walking aid, rather than the patient's underlying motion mechanics~\cite{zhang2024enhancing,Geirhos2020ShortcutLearning,McArdle2021Environment}.

Recently, Large Vision-Language Models (LVLMs)~\cite{zhu2025internvl3exploringadvancedtraining,liu2023llava,li2023llava,ji2026biocoach} offer a potential solution to this interpretability gap. However, standard LVLMs primarily rely on static visual cues and lack the domain knowledge to quantify fine-grained temporal abnormalities (\emph{e.g.}, foot drop). Furthermore, when deployed on uncontrolled ``in-the-wild'' videos, they are prone to \textit{shortcut learning}: attending to environmental confounders (\emph{e.g.}, a hospital bed) rather than the patient's underlying motion mechanics~\cite{zhang2024enhancing,Geirhos2020ShortcutLearning,McArdle2021Environment}.

To address both robustness and clinical interpretability, we propose \textbf{BioGait-VLM}, a tri-modal framework that seamlessly integrates RGB vision, natural language, and explicit biomechanics. Conceptually, our approach mitigates the reliance on superficial visual shortcuts by enforcing a dual-path reasoning process: while the visual stream captures contextual appearance, a dedicated biomechanical stream isolates pure kinematic dynamics. 
Crucially, this design supports a \emph{\textbf{dual-output capability:}} the fused representations drive a specialized classification head for high-precision diagnosis, while the underlying language model retains the generative capacity to output human-readable clinical descriptions, providing transparent reasoning for its predictions.

%%
% Structurally, BioGait-VLM creates a holistic clinical representation by bridging the modality gap between pixel-level features and high-level medical reasoning. While we leverage a pre-trained Visual Branch to extract semantic appearance cues, standard VLMs lack temporal sensitivity. To address this, we integrate a specialized Temporal Evidence Distillation Branch that compresses frame-level features into dynamic gait descriptors, allowing the model to detect subtle temporal pathologies like festination or tremors. Uniquely, rather than using a generic skeleton encoder, our Biomechanical Branch employs a novel tokenization strategy that projects 3D kinematics into language-aligned semantic tokens, enabling the LLM to explicitly ``read'' the patient's motion. 
% %
% To rigorously validate this architecture, we extend the public GAVD benchmark~\cite{Ranjan2025GAVD} by integrating our collected Degenerative Cervical Myelopathy (DCM) Dataset, a high-fidelity clinical cohort of 30 patients, resulting in a unified $8$-class pathological taxonomy comprising a total of $1,181$ video clips. We rectify this combined benchmark with leakage-aware subject-disjoint splits, ensuring that clips from the same patient never overlap between training and testing. By evaluating on this challenging multi-class setting and conducting a blinded human-expert study, we demonstrate that our tri-modal approach not only improves generalization but also generates clinically grounded, trustworthy assessments.

Structurally, BioGait-VLM creates a holistic clinical representation by bridging pixel-level features and medical reasoning. While we leverage a pre-trained Visual Branch for appearance cues, standard VLMs lack temporal sensitivity. To address this, we integrate a Temporal Evidence Distillation Branch that compresses frame features into dynamic gait descriptors, allowing the model to detect subtle pathologies like festination. Uniquely, our Biomechanical Branch employs a novel tokenization strategy that projects 3D kinematics into language-aligned semantic tokens, enabling the LLM to explicitly ``read'' the patient's motion.
To rigorously validate this architecture, we extend the public GAVD benchmark~\cite{Ranjan2025GAVD} by integrating our collected Degenerative Cervical Myelopathy (DCM) Dataset, a high-fidelity clinical cohort of 30 patients, resulting in a unified $8$-class pathological taxonomy comprising a total of $1,181$ video clips. We construct this combined benchmark with leakage-aware subject-disjoint splits, ensuring that clips from the same patient never overlap between training and testing.
By evaluating on this challenging setting and conducting a blinded expert study, we demonstrate that our tri-modal approach achieves state-of-the-art generalization and generates clinically grounded assessment.

In summary, the contributions of this work include:

$\diamond$ We propose \textbf{BioGait-VLM}, a novel tri-modal framework that introduces Biomechanical Tokenization, projecting 3D kinematics into language-aligned tokens to enable robust reasoning and interpretable clinical assessment.

$\diamond$ %We integrate a specialized \emph{Temporal Evidence Distillation Branch} that compresses frame-level features into dynamic descriptors, capturing fine-grained temporal pathologies often missed by standard Vision-Language Models
We integrate a specialized \emph{Temporal Evidence Distillation Branch} that compresses frame-level features into compact dynamic descriptors, ensuring the model captures temporal evolution rather than just static appearance.

$\diamond$ We collect and curate the DCM Clinical Dataset, a high-fidelity, expert-validated benchmark for Degenerative Cervical Myelopathy, enabling rigorous evaluation of fine-grained pathological motion in authentic clinical settings.

$\diamond$ We achieve state-of-the-art performance on clinical benchmarks and confirm via a human-expert study that our approach significantly improves clinical correctness and evidence grounding.

%
%\input{sec_arXiv/2_related_work}
%

%--------------------------------------------------------------%
\section{Methodology}
\label{sec:method}

%--------------------------------------------------------------%
\subsection{Problem Formulation and Overview}

We formulate video-based clinical gait analysis as a dual-objective task: \textit{classification} and \textit{interpretable reasoning}.
Given a gait video $\mathcal{V} = \{\mathbf{x}_t\}_{t=1}^T$ and an associated 3D skeleton sequence $\mathcal{S}$ (inferred via HSMR~\cite{Xia2025HSMR}), the model aims to predict a pathology label $y \in \{1, \dots, K\}$ (where $K{=}8$ in our experiments, includes \textit{Abnormal}, \textit{Myopathic}, \textit{Parkinson's}, etc.) while simultaneously generating a textual rationale $\mathcal{T}$ describing the biomechanical evidence.

\begin{figure}[t]
\centering
\includegraphics[width=0.98\textwidth]{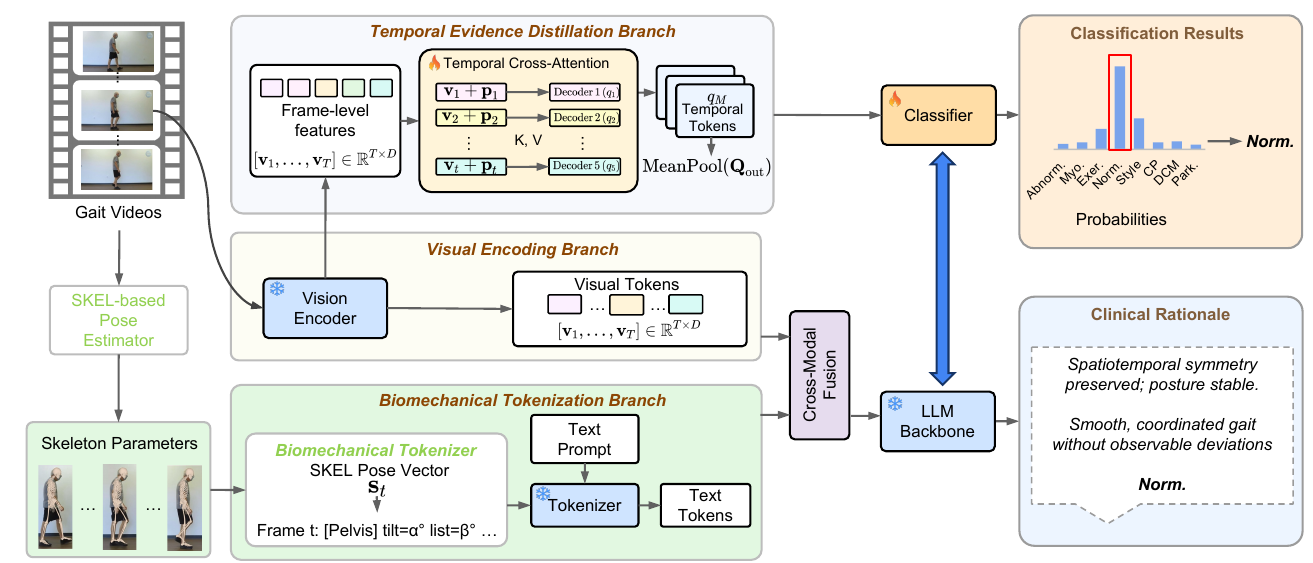}
%\vspace{-2mm}
%\caption{\small \textbf{Overview of BioGait-VLM framework}. Our tri-modal architecture disentangles gait into three streams: (Top) A new Temporal Evidence Distillation (TED) branch to recover fine-grained rhythmic dynamics (\emph{e.g.}, tremors); (Middle) Visual Context for appearance; and (Bottom) A novel Biomechanical Tokenization branch that projects 3D kinematics into language-aligned semantic tokens. These modalities are fused in a frozen LVLM to yield state-of-the-art classification and interpretable, evidence-grounded clinical reports.}
%\vspace{-4mm}
\caption{\textbf{Overview of BioGait-VLM.} The framework integrates three streams: (Top) Temporal Evidence Distillation (TED) for rhythmic dynamics; (Middle) Visual Context; and (Bottom) Biomechanical Tokenization, projecting 3D kinematics into semantic tokens. Fused within a frozen LVLM, these modalities enable state-of-the-art classification and evidence-grounded clinical reporting.}
%\vspace{-2mm}
\label{fig:fig1}
%\vspace{-3mm}
\end{figure}

%%
%To achieve this, \textbf{BioGait-VLM} adopts a \textbf{tri-branch architecture} (Fig.~\ref{fig:fig1}) that fuses three complementary information pathways:
To achieve this, we construct \textbf{BioGait-VLM} upon InternVL~\cite{zhu2025internvl3exploringadvancedtraining}, a pre-trained LVLM. We select this foundation to leverage its \emph{pre-aligned semantic space}, which naturally accommodates our textualized biomechanical tokens, and its generative capabilities for clinical reporting. Furthermore, by freezing the massive pre-trained parameters and training only auxiliary modules, we ensure robust generalization even on limited clinical datasets. As illustrated in Fig.~\ref{fig:fig1}, the framework processes inputs through three complementary pathways:

\begin{enumerate}
    \item \textbf{Visual Encoding Branch}: A frozen vision encoder extracts spatial features from $\mathcal{V}$ to capture contextual appearance.
    \item \textbf{Temporal Evidence Distillation (TED) Branch}: A query-based decoder aggregates frame-level features into compact gait descriptors, explicitly modeling \emph{temporal dynamics}.
    \item \textbf{Biomechanical Tokenization Branch}: A kinematic tokenizer projects the skeleton sequence $\mathcal{S}$ into language-aligned semantic tokens, enabling the model to reason about joint mechanics independent of visual appearance.
\end{enumerate}

%-------------------- figure 1 -----------------------
%-------------------- figure 1 -----------------------

%--------------------------------------------------------------%
\subsection{Tri-Modal Vision-Language Architecture}
\paragraph{Visual Encoding Branch.}
Given $T$ sampled frames ($T{=}32$ in our experiment), we employ the frozen InternVL vision encoder $f_\text{vis}$ to process each frame $\mathbf{x}_t$ independently. Unlike standard action recognition backbones that output abstract feature maps, InternVL yields semantically rich visual tokens. We perform spatial mean-pooling on the patch embeddings to obtain a sequence of frame-level visual features $\mathbf{V}$:
$\mathbf{V} = [\mathbf{v}_1, \dots, \mathbf{v}_T] \in \mathbb{R}^{T \times D}$,
where $\mathbf{v}_t = \text{MeanPool}(f_\text{vis}(\mathbf{x}_t))$ and $D$ denotes the hidden dimension of the language model.

%--------------------------------------------------------------%
\paragraph{Temporal Evidence Distillation (TED) Branch.}
Standard VLMs typically process video via naive mean-pooling, which can obscure fine-grained rhythmic cues essential for gait diagnosis, such as tremors or festination. To recover these dynamics, we propose a new Temporal Evidence Distillation module.

Instead of simple pooling, we initialize $M{=}32$ learnable motion queries $\mathbf{Q} \in \mathbb{R}^{M \times D}$. These queries function as latent temporal anchors, learning to attend to specific phases of the gait cycle across the visual feature sequence $\mathbf{V}$. To preserve sequential order, we inject learnable temporal positional embeddings $\mathbf{P} \in \mathbb{R}^{T \times D}$:
\begin{equation}
    \mathbf{Q}_\text{out} = \text{TransformerDecoder}(\mathbf{Q},\; \mathbf{V} + \mathbf{P}).
\end{equation}
The decoder comprises $L{=}3$ layers with $H{=}4$ attention heads. Through cross-attention, the queries dynamically aggregate frame-level features based on their relevance to motion patterns rather than visual appearance. 
%The outputs are condensed into a single dynamic gait descriptor via mean-pooling:
We employ mean-pooling on the output queries to form a \emph{global dynamic descriptor} that captures the presence of rhythmic anomalies across the sequence, regardless of their specific timestamp:
    $\mathbf{f}_\text{temp} = \text{MeanPool}(\mathbf{Q}_\text{out}) \in \mathbb{R}^{D}$.
This explicitly preserves high-frequency motion anomalies independent of the VLM's semantic context.
%This branch ensures that high-frequency motion anomalies are preserved even when the primary VLM backbone focuses on high-level semantic context.

%--------------------------------------------------------------%
\paragraph{Biomechanical Tokenization Branch.}
To disentangle pathological motion from environmental confounders, we introduce a strategy to project kinematics into the semantic space of the LLM. First, we extract a 46-dimensional SKEL pose vector $\mathbf{s}_t$ for each frame from the skeleton sequence $\mathcal{S}$ using the off-the-shelf HSMR estimator~\cite{Keller2023SKEL,Xia2025HSMR}. This covers clinically critical joints including the pelvis, spine, and limbs.
We convert these numerical tensors into structured natural-language descriptions using a template-based mapping function. For a given frame $t$, the joint angles are formatted as:
\begin{quote}
\small\texttt{Frame $t$: [Pelvis] tilt=$\alpha$° list=$\beta$° ... [R.Knee] flex=$\delta$°}
\end{quote}
The descriptions for all $T$ frames are concatenated with a task-specific clinical instruction prompt and tokenized to produce a sequence of biomechanical embedding tokens $\mathbf{E}_\text{bio}$:
    $\mathbf{E}_\text{bio} = [\mathbf{e}_1, \dots, \mathbf{e}_{L_\text{text}}] \in \mathbb{R}^{L_\text{text} \times D}$.
This approach eliminates the need for a separate skeleton encoder and aligns the kinematic data with the pre-trained semantic knowledge of the LLM, effectively allowing the model to ``read'' the patient's motion.

%--------------------------------------------------------------%
\paragraph{Cross-Modal Fusion}
We concatenate visual features $\mathbf{V}$ and biomechanical tokens $\mathbf{E}_\text{bio}$ along the sequence dimension:
\begin{equation}
    \mathbf{Z}_\text{input} = [\mathbf{V}; \mathbf{E}_\text{bio}] \in \mathbb{R}^{(T + L_\text{text}) \times D}.
\end{equation}
The frozen LLM processes the sequence. We pool visual hidden states into $\mathbf{f}_\text{vlm}$ and concatenate this with the explicit temporal descriptor $\mathbf{f}_\text{temp}$ to form the holistic representation $\mathbf{f}_\text{final} \in \mathbb{R}^{2D}$.

%This sequence is processed by the frozen LLM. We extract and pool the last hidden states corresponding to visual positions to obtain $\mathbf{f}_\text{vlm}$. Finally, we concatenate this with the explicit temporal descriptor $\mathbf{f}_\text{temp}$ to form the holistic representation $\mathbf{f}_\text{final} \in \mathbb{R}^{2D}$.

%--------------------------------------------------------------%
\subsection{Clinical Instruction Tuning and Objectives}
\label{sec:training}

\paragraph{Training Objective.}
To mitigate class imbalance, we use weighted cross-entropy loss. A linear head $g: \mathbb{R}^{2D} \rightarrow \mathbb{R}^{K}$ maps $\mathbf{f}_\text{final}$ to logits:
\begin{equation}
    \mathcal{L} = -\sum_{k=1}^{K} w_k\, y_k \log \frac{\exp(g(\mathbf{f}_\text{final})_k)}{\sum_{j} \exp(g(\mathbf{f}_\text{final})_j)}, \quad w_k = \frac{N}{K \cdot n_k}.
\end{equation}
We update only the Temporal Decoder and linear head, freezing InternVL.
%We update only the Temporal Decoder and linear head, freezing InternVL.

\paragraph{Evidence-Grounded Prompting.}
We employ a \textit{Clinical Instruction Prompt} that defines each gait class using kinematic descriptors (\emph{e.g.}, \textit{``Parkinsonian: shuffling steps, reduced arm swing''}). Injecting these definitions alongside $\mathbf{E}_\text{bio}$ primes the model to attend to specific joint anomalies during the forward pass.

\subsection{The DCM Clinical Gait Dataset}
\label{sec:dataset}

To bridge the gap between uncurated web videos and clinical reality, we introduce the Degenerative Cervical Myelopathy (DCM) Dataset, a high-fidelity cohort collected during routine neurosurgical assessments. Unlike public datasets (\emph{i.e.}, GAVD) which often lack specific spinal pathologies, our dataset specifically targets DCM, a condition characterized by subtle, often misdiagnosed gait anomalies such as spasticity, imbalance, and increased cadence.

\begin{figure}[t]
\centering
\includegraphics[width=0.95\textwidth]{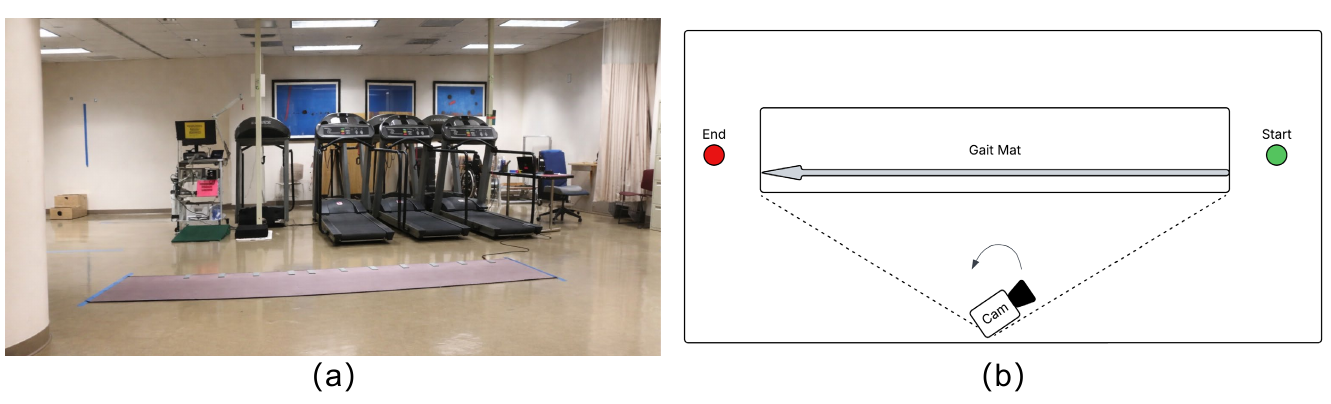}
\caption{\textbf{DCM Dataset Acquisition Setup.} (a) The recording environment at the outpatient clinic. The scene includes realistic background clutter (\emph{e.g.}, treadmills, medical equipment) rather than a sterile lab background, providing a rigorous testbed for evaluating model robustness against visual shortcuts. (b) Top-down schematic of the standardized capture protocol. A tripod-mounted RGB camera is positioned to capture the sagittal (side) view of the patient traversing the instrumented gait mat, ensuring consistent biomechanical visibility.}
\label{fig:dataset}
\end{figure}

\paragraph{Acquisition Protocol.}
Patients were recruited during outpatient neurosurgical visits from elective spine surgery patients treated at a tertiary care center in Washington University School of Medicine. Participation was voluntary, with informed consent obtained prior to data collection. The dataset comprises $239$ video recordings from $30$ unique patients; recruitment remains active as part of an ongoing clinical study to further expand the cohort's scale and diversity. As illustrated in Fig.~\ref{fig:dataset}, trials were recorded using a floor-mat gait assessment setup with synchronized RGB video capture (30 fps, 1080p). We utilized a standardized sagittal-view protocol along a 10-meter hospital corridor. This setup ensures a consistent viewing angle while retaining realistic clinical background clutter (\emph{e.g.}, medical equipment, flooring patterns), providing a rigorous testbed for evaluating model robustness against environmental shortcuts.

%-------------------- dataset -----------------------
%-------------------- dataset -----------------------

\paragraph{Expert Annotation and Integration.}
Ground truth labels were established based on the diagnosis from an attending spine surgeon. Each patient was assessed for specific biomechanical markers of myelopathy. Patients were also given DCM-related questionnaires for daily symptoms. We integrate these high-quality samples into the public GAVD benchmark to form our unified 8-class taxonomy. The DCM samples specifically enrich the underrepresented pathological categories, providing fine-grained examples distinct from generic ``abnormal'' gait.

\paragraph{Privacy and Ethical Compliance.}
Data collection adhered to strict HIPAA and IRB guidelines. Raw videos were processed on a secure computing instance at Washington University by approved personnel. To ensure privacy, we implemented a dual-stream de-identification pipeline: (1) a Skeleton-Processing Path, where kinematic parameters were extracted using HSMR without retaining identifiers; and (2) a Visual-Embedding Path, where face blurring was applied prior to feature extraction. Only these de-identified derivatives were used for model training, ensuring that no protected health information (PHI) is exposed in the pipeline. All data were stored on HIPAA-compliant servers. All computer vision analyses involving identifiable data were performed on HIPAA-compliant local servers. 

% \subsection{The DCM Clinical Gait Dataset}
% To bridge the gap between uncurated web videos and clinical reality, we introduce the Degenerative Cervical Myelopathy (DCM) Dataset, a high-fidelity cohort comprising $239$ video recordings from $30$ patients collected during routine neurosurgical assessments. Unlike variable internet clips, these videos are captured using a standardized sagittal-view protocol in a hospital environment, yet retain realistic background clutter to rigorously test model robustness against environmental shortcuts. Ground truth labels are established via consensus among senior neurosurgeons and physiotherapists, specifically enriching the underrepresented \textit{Myopathic} and \textit{Abnormal} categories within our unified 8-class taxonomy. All data collection is conducted under Institutional Review Board (IRB) approval with strict de-identification protocols, establishing a trustworthy benchmark for fine-grained pathological gait analysis.

\subsection{Implementation Details}
We implement the framework using InternVL3.5-1B~\cite{zhu2025internvl3exploringadvancedtraining} as the backbone, with input frames resized to $448{\times}448$. The Temporal Evidence Distillation decoder is configured with $L{=}3$ layers and hidden dimension $D{=}2048$. Biomechanical text sequences are truncated to a maximum length of 1024 tokens. Model training is conducted on a single NVIDIA A100 GPU using the AdamW optimizer ($lr{=}5{\times}10^{-4}$, batch size 2) for 40 epochs.

%

%--------------------------------------------------------------%
\section{Experiments} 
\label{sec:experiments}

\subsection{Experimental Settings}
We evaluate on a unified clinical gait benchmark spanning an 8-class taxonomy: \textit{Abnormal} (Abnorm.), \textit{Normal} (Norm.), \textit{Style}, \textit{Exercise}, \textit{Parkinson's} (Park.), \textit{Cerebral Palsy (CP)}, \textit{Myopathic} (Myo.), and our newly introduced \textit{DCM}.

\paragraph{GAVD~\cite{Ranjan2025GAVD} (Public Component).} 
We utilize the Gait Abnormality Video Dataset to source established gait patterns. This includes \textit{Normal} controls and broad \textit{Abnormal} gait cases, non-pathological variations (\textit{Style}, \textit{Exercise}), and standard neuropathologies (\textit{Parkinson's}, \textit{CP}, \textit{Myopathic}).

\paragraph{DCM (Ours).} 
To address the lack of spinal cord pathologies in public data, we integrate our newly collected Degenerative Cervical Myelopathy (DCM) dataset. In our 8-class benchmark, these samples constitute the distinct \emph{DCM} class, representing a subtle pathological motion often confused with generic abnormalities.
While the cohort size (30 patients) reflects the scarcity of annotated pathological gait data compared to general action recognition, the high-fidelity expert validation ensures that evaluation on the held-out test set (52 clips) represents a dense, clinically significant signal.

% \paragraph{GAVD~\cite{Ranjan2025GAVD} (Public Component).} 
% We utilize the Gait Abnormality Video Dataset~\cite{Ranjan2025GAVD}, re-partitioned to strictly enforce subject independence. This subset comprises $942$ sequences from $198$ subjects. We allocate $158$ subjects ($728$ sequences) to training and $40$ subjects ($214$ sequences) to testing.

% \paragraph{DCM (Ours).} 
% To address the lack of spinal cord pathologies, we integrate our collected Degenerative Cervical Myelopathy dataset. This component consists of $239$ sequences from $30$ patients. Adhering to an 8:2 subject-level split, we assign $24$ patients ($187$ sequences) to the training set and hold out $6$ patients ($52$ sequences) for testing to assess generalization to unseen clinical cases.

\paragraph{Leakage-Aware Protocol.} 
% To prevent the model from memorizing patient identities, we enforce \textbf{subject-disjoint splits}. For GAVD, clips from the same source video are kept in the same partition. For DCM, splits are stratified by Patient ID. The combined dataset comprises 915 training and 266 testing sequences, ensuring a rigorous test of generalization to unseen subjects.
%
 Unlike~\cite{Ranjan2025GAVD}, to prevent the model from memorizing patient identities, we re-partition the benchmark to strictly enforce subject-disjoint splits. For GAVD, we allocate $158$ subjects ($728$ sequences) to training and hold out $40$ subjects ($214$ sequences) for testing. For DCM, we adhere to an $8:2$ patient-level split, assigning $24$ patients ($187$ sequences) to training and $6$ patients ($52$ sequences) to testing. The final combined benchmark comprises $915$ training and $266$ testing sequences, ensuring zero subject overlap.

\paragraph{Baselines.}
We compare BioGait-VLM against two categories of methods:
(1) \emph{End-to-End Video Encoders}: 
We re-train standard 3D-CNN and Transformer architectures, specifically SlowFast~\cite{Feichtenhofer2019SlowFast} and TSN~\cite{WangTSNTPAMI2019}, on our leakage-aware splits. These represent ``black-box'' approaches that learn abstract spatiotemporal features from pixels without explicit biomechanics.
(2) \emph{General-Purpose LVLMs}: We evaluate two SoTA foundational large vision-language models: Qwen3-VL~\cite{bai2025qwen3} and InternVL3.5~\cite{zhu2025internvl3exploringadvancedtraining}, in a zero-shot setting to assess whether general pre-training suffices for fine-grained clinical diagnosis.

%We fine-tune widely used backbones, specifically SlowFast~\cite{Feichtenhofer2019SlowFast} and TSN~\cite{WangTSNTPAMI2019}, to assess the performance of pure pixel-based kinematic learning.
%(2) Large Vision-Language Models (LVLMs): We evaluate general-purpose foundation models, including Qwen2-VL-2B~\cite{qwen3vl} and InternVL3.5-1B~\cite{zhu2025internvl3exploringadvancedtraining}, in a zero-shot setting to determine if general pre-training is sufficient for fine-grained clinical diagnosis.

%--------------------------------------------------------------%

\begin{table}[t]
\centering
\caption{\small \textbf{Results \& Ablation.} Performance on the $8$-class clinical gait benchmark. Top: Comparison against state-of-the-art video and VLM baselines. Bottom: Ablation study isolating the impact of the \emph{Temporal Evidence Distillation} (TED) and \emph{Biomechanical Tokenization} branches.  %\textit{Note that the Full BioGait-VLM} achieves the best balance across diverse pathologies.
\textbf{Note:} Test set class distribution is imbalanced; consequently, F1 scores on underrepresented classes may exhibit higher variance.
}
%\caption{\small \textbf{Main Results.} Classification performance (Accuracy and F1) on the unified \textbf{8-class} clinical gait benchmark. \textbf{Bold} indicates best performance. Note that while baselines struggle with subtle pathologies (DCM, Myopathic), BioGait-VLM achieves consistent gains across all clinical categories.}
\label{tab:sota}
\resizebox{\textwidth}{!}{%
\begin{tabular}{l|c|c|cccccccc}
%\rowcolor{tabHeader2}
\toprule
\multirow{2}{*}{\textbf{Method}} &  \multirow{2}{*}{\textbf{Acc. (\%)}} &  \multirow{2}{*}{\textbf{Macro F1}} & \multicolumn{8}{c}{\textbf{Per-Class F1 Score (\%)}} \\
\cline{4-11}
 &  & & DCM & Myo. & Abnorm. & CP & Park. & Norm. & Style & Exer. \\
 \rowcolor{tabHeader2}
\midrule
\multicolumn{11}{l}{\textit{End-to-End Video Encoders (Fine-tuned)}} \\
SlowFast~\cite{Feichtenhofer2019SlowFast} & $32.2$ & $31.1$ & $88.2$ & $54.1$ & $21.5$ & $0.0$ & $57.1$ & $0.0$ & $3.9$ & $24.3$ \\
TSN~\cite{WangTSNTPAMI2019} & $37.1$ & $35.9$ & $100.0$ & $47.6$ & $23.4$ & $12.1$ & $75.0$ & $0.0$ & $0.0$  & $29.2$  \\
\rowcolor{tabHeader2}
\midrule
\multicolumn{11}{l}{\textit{General-Purpose LVLMs (Zero-Shot)}} \\
Qwen3-VL-2B~\cite{bai2025qwen3} & $21.9$ & $7.8$ & $0.0$ & $0.0$ & $46.7$ & $0.0$ & $0.0$ & $15.7$ & $0.0$ & $0.0$ \\
InternVL3.5-1B~\cite{zhu2025internvl3exploringadvancedtraining} & $32.5$ & $9.6$ & $0.0$ & $0.0$ & $52.3$ & $0.0$ & $0.0$ & $7.6$ & $8.0$ & $9.2$ \\
\rowcolor{ourbg}
\midrule
\multicolumn{11}{l}{\textit{BioGait-VLM Ablation Study (Component Analysis)}} \\
 \, - \textit{w/o} TED Branch  & $63.8$ & $42.4$ & $93.3$ & $35.5$ & $71.9$ & $10.0$ & $64.1$ & $0.0$ & $4.2$ & $60.2$ \\
\, - \textit{w/o} Biomechanical Branch  & $64.3$ & $43.8$ & $98.1$ & $33.3$ & $73.4$ & $16.0$ & $65.7$ & $0.0$ & $0.0$ & $64.4$ \\
\, - \textit{w/o} both & $60.0$ & $44.2$ & $\textbf{100.0}$ & $30.0$ & $72.9$ & $0.0$ & $66.7$ & $21.1$ & $0.0$ & $63.3$ \\
\rowcolor{gbg}
\midrule
\textbf{BioGait-VLM} (Full) & $\textbf{68.1}$ & $\textbf{52.9}$ & $98.1$ & $\textbf{36.4}$ & $\textbf{80.4}$ & $\textbf{35.3}$ & $\textbf{66.7}$ & $\textbf{31.8}$ & $\textbf{4.3}$ & $\textbf{70.0}$ \\
\bottomrule
\end{tabular}%
}
\end{table}

%-------------------- vis  -----------------------
\begin{figure}[ht!]
\centering
\includegraphics[width=0.98\textwidth]{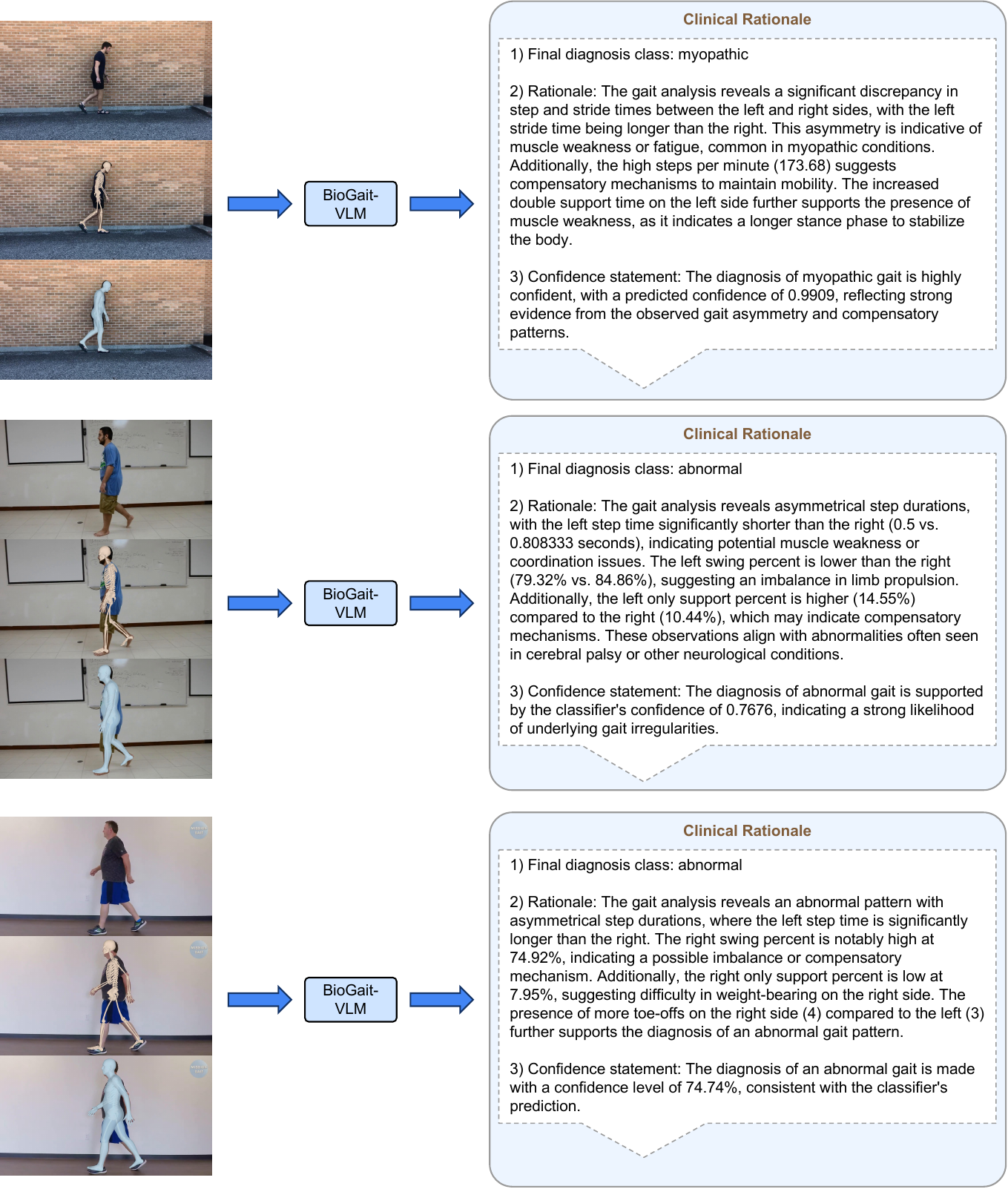}
\caption{\textbf{Qualitative Visualization of Interpretable Reasoning.} Our BioGait-VLM generates detailed clinical rationales anchored in quantitative evidence.
\emph{(Top) Myopathic:} The model identifies ``muscle weakness'' via specific metrics, citing ``high steps per minute (173.68)'' and ``increased double support time.''
\emph{(Middle) Abnormal:} It detects potential coordination issues by calculating precise asymmetry: ``left step time significantly shorter (0.5s vs. 0.81s)'' and ``imbalance in limb propulsion.''
\emph{(Bottom) Abnormal:} It distinguishes a different irregularity by counting specific events, noting ``more toe-offs on the right side (4) compared to the left (3)'' and quantifying swing percentage.
These examples demonstrate how \emph{Biomechanical Tokenization} enables granular, evidence-based differentiation even within broad diagnostic categories.}
\label{fig:vis}
\end{figure}
%-------------------- vis -----------------------

%-------------------------------------------
\subsection{Diagnostic Classification Performance}
Tab~\ref{tab:sota} (top) presents results on the unified 8-class benchmark. Standard video encoders (SlowFast, TSN) and zero-shot VLMs struggle to generalize under the rigorous subject-disjoint protocol, achieving accuracies below 40\%. This highlights their reliance on environmental shortcuts and inability to capture fine-grained kinematic deviations.
%
%In contrast, \textbf{BioGait-VLM} achieves a state-of-the-art accuracy of \textbf{\emph{68.1\%}}, outperforming the strongest baseline by $+\textbf{\emph{31.0}}\%$. Crucially, our framework delivers substantial gains on subtle pathologies, achieving F1 scores of \textbf{\emph{98.1\%}} for \emph{DCM} and \textbf{\emph{66.7\%}} for \emph{Parkinson's}. This confirms that explicitly modeling biomechanics and temporal dynamics enables the system to disentangle pathological motion from complex environmental noise.
%In contrast, \textbf{BioGait-VLM} achieves state-of-the-art performance, with \textbf{\emph{68.1\%}} accuracy and \textbf{52.9\%} Macro F1, outperforming the strongest baseline (TSN) by \textbf{\emph{+17.0 pp}} in Macro F1. Crucially, BioGait-VLM yields large gains on clinically challenging categories where prior methods struggle: it attains \textbf{\emph{80.4\%}} F1 on \emph{Abnormal gait} (vs.\ 23.4\% for TSN). This diagnostic precision is supported by the model's interpretable reasoning (Fig.~\ref{fig:vis}). By grounding predictions in explicit kinematic deviations, \emph{e.g.}, ``imbalance in limb propulsion'' or ``toe-off asymmetry'', our BioGait-VLM can distinguish generic abnormalities from specific pathologies, enabling robust classification even when visual cues are ambiguous.
%
In contrast, \textbf{BioGait-VLM} achieves state-of-the-art performance with \textbf{\emph{68.1\%}} accuracy. While we report a Macro F1 of \textbf{\emph{52.9\%}}, we note that this metric can be sensitive to variance in underrepresented test classes. Crucially, on pathologically significant categories with robust evaluation data, our framework delivers massive gains: it attains \textbf{\emph{98.1\%}} F1 on \emph{DCM} and \textbf{\emph{80.4\%}} on \emph{Abnormal gait} (vs.\ 23.4\% for TSN). This diagnostic precision is underpinned by the model's interpretable reasoning (Fig.~\ref{fig:vis}). By grounding predictions in explicit kinematic deviations, BioGait-VLM successfully disentangles specific pathologies from generic abnormalities, enabling robust classification even when visual cues are ambiguous.

%------------%
\subsection{Clinical Validation via Blinded Expert Assessment}
\label{sec:human_study}

Quantitative metrics (accuracy/F1) do not fully capture a model's utility in a clinical workflow. To assess the interpretability and trustworthiness of the generated rationales, we conduct a \emph{blinded qualitative study} with four evaluators possessing specific domain expertise in DCM.

\begin{figure}[!htbp]
\centering
\includegraphics[width=0.76\textwidth]{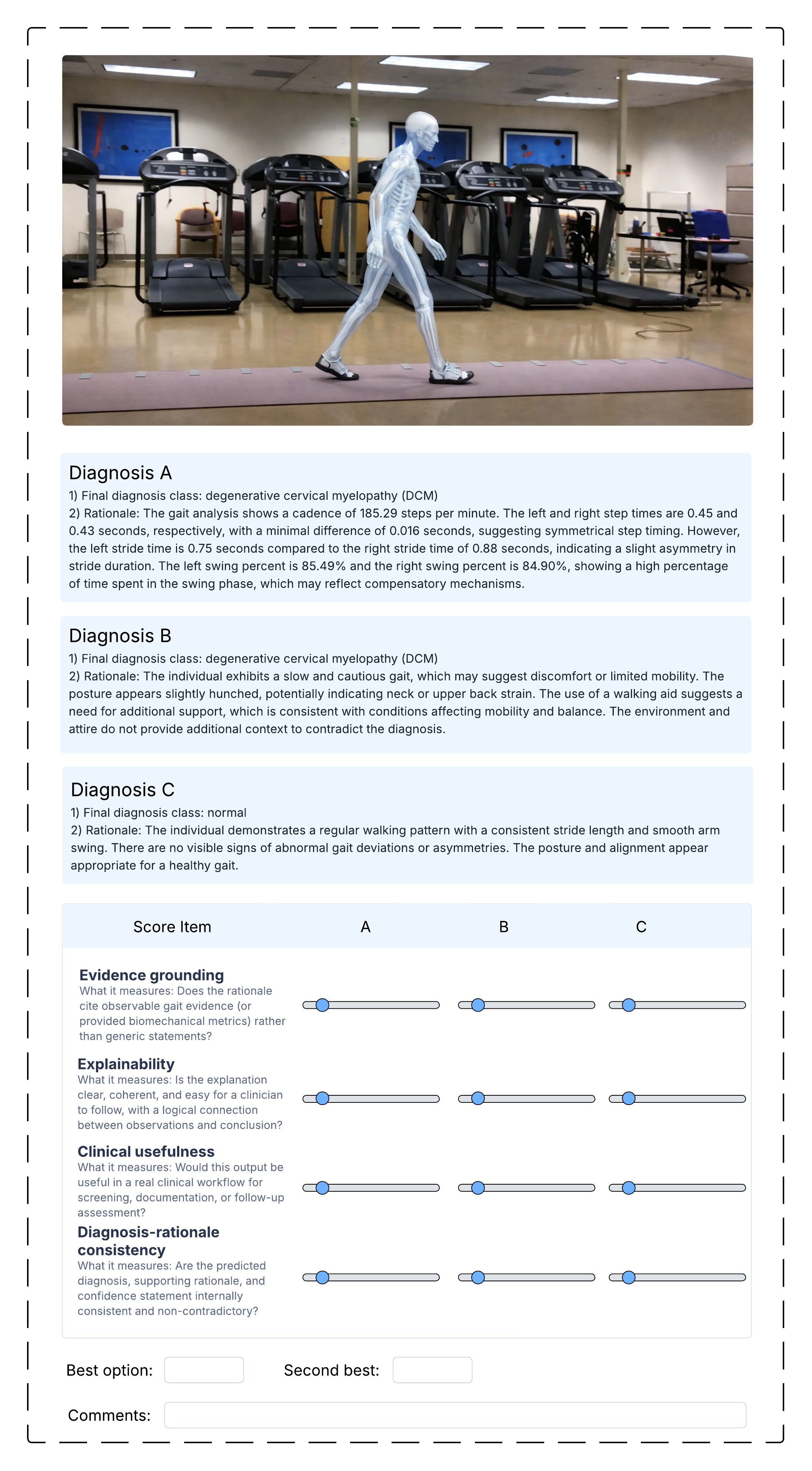}
\vspace{-4mm}
\caption{\textbf{Interface for Blinded Expert Evaluation.} To ensure unbiased assessment, we developed a custom web-based annotation tool. Clinicians view the patient gait video alongside three anonymized and randomized model-generated reports (labeled Diagnosis A, B, C). Experts rate each rationale independently on four clinical dimensions using Likert scales before selecting the superior model, ensuring that rankings are based solely on clinical quality rather than model identity.}
\label{fig:interface}
\end{figure}

\paragraph{Study Protocol.}
We utilize the complete set of 52 video sequences derived from all 6 patients in the held-out DCM test cohort. To ensure exhaustive coverage without fatigue, the dataset is partitioned into non-overlapping subsets, with each evaluator reviewing 13 unique cases. For every sequence, 
 experts reviewed diagnostic rationales using a custom-developed blinded interface (Fig.~\ref{fig:interface}).
The platform randomized the order of the following three \textbf{\emph{anonymized}} models as 'Diagnosis A/B/C' to prevent selection bias:
(1) \emph{InternVL-3.5 (Zero-Shot):} The generic backbone without adaptation.
(2) \emph{BioGait-VLM (w/o biomechanical branch):} fine-tuned model without the biomechanical branch.
(3) \emph{BioGait-VLM (Full):} The proposed framework.
Evaluators rate the responses on a 5-point Likert scale across four clinical dimensions:

 \begin{center}
	\vspace*{-0.5cm}
	\setlength\fboxrule{0.0pt}
	\noindent\fcolorbox{black}[rgb]{0.95,0.95,0.95}{\begin{minipage}{1\columnwidth}
\begin{enumerate}
    \item \textbf{Evidence Grounding:} Does the rationale cite observable biomechanical metrics (\emph{e.g.}, ``reduced knee flexion'') rather than generic statements?
    \item \textbf{Explainability:} Is the explanation coherent, logical, and easy for a clinician to follow?
    \item \textbf{Clinical Usefulness:} Would this output aid real-world screening or documentation?
    \item \textbf{Consistency:} Is the predicted diagnosis internally consistent with the supporting text?
\end{enumerate}
	\end{minipage}}
	\vspace*{-0.3cm}
\end{center}

% \begin{enumerate}
%     \item \textbf{Evidence Grounding:} Does the rationale cite observable biomechanical metrics (\emph{e.g.}, ``reduced knee flexion'') rather than generic statements?
%     \item \textbf{Explainability:} Is the explanation coherent, logical, and easy for a clinician to follow?
%     \item \textbf{Clinical Usefulness:} Would this output aid real-world screening or documentation?
%     \item \textbf{Consistency:} Is the predicted diagnosis internally consistent with the supporting text?
% \end{enumerate}

Finally, experts select the single ``Best Model'' for each case and provide open-ended comments to justify their selection.
\textbf{Note:} While this study is designed as a focused qualitative pilot, the use of high-level domain experts ensures that the assessments reflect rigorous clinical standards.

\begin{table}[t]
\centering
\caption{\textbf{Human Expert Evaluation.} Average likert scores (Range: 1--5, higher is better). Our \emph{Biomechanical Tokenization Branch} significantly boosts Evidence Grounding and Usefulness compared to vision-only approaches, reducing hallucinations.}
\label{tab:human_study}
\resizebox{0.98\columnwidth}{!}{%
\begin{tabular}{l|cccc}
\toprule
\textbf{Model Variant} & \textbf{Grounding} & \textbf{Explainability} & \textbf{Usefulness} & \textbf{Consistency} \\
\midrule
InternVL3.5-1B~\cite{zhu2025internvl3exploringadvancedtraining} (Zero-Shot) & $2.31$ & $2.98$ & $2.46$ & $3.85$ \\
\textbf{BioGait-VLM} (\emph{w/o Biomechanical Branch}) & $1.98$ & $2.12$ & $1.83$ & $2.13$ \\
\rowcolor{gbg}
\textbf{BioGait-VLM} (Full) & $\textbf{4.23}$ & $\textbf{3.98}$ & $\textbf{3.69}$ & $\textbf{4.15}$ \\
\bottomrule
\end{tabular}%
}
\end{table}

%-------------------- human study  -----------------------
%-------------------- human study -----------------------

\paragraph{Results and Analysis.}
As detailed in Tab.~\ref{tab:human_study}, BioGait-VLM dominates across all metrics, particularly in Evidence Grounding ($\textbf{\emph{4.23}}$ vs. $2.31$ for the strongest baseline). In the comparative ranking, experts select BioGait-VLM as the superior model in $\textbf{\emph{69.2\%}}$ of cases (36/52), a statistically significant preference over the baselines (\textbf{$p < 0.001$}, binomial test).
Qualitative feedback highlights the critical role of the Biomechanical Tokenization Branch. Evaluators noted that while baseline models often provided ``generic'' or ``vague'' explanations (\emph{e.g.}, ``walking difficulty''), the Full Model produced rationales anchored in ``concrete metrics.'' One expert commented: \textit{``Even though [the baseline] makes sense, [BioGait-VLM] wins with concrete metrics.''} Another noted that even when visual cues were ambiguous, \textit{``the biomechanical metrics present a better rationale... the data suggests otherwise.''} This confirms that textualized kinematics transform the system from a black-box classifier into a transparent, evidence-based clinical assistant.

%consistently outperforms baselines, particularly in \emph{Evidence Grounding}.Qualitative feedback indicates that the \textit{Zero-Shot} baseline frequently hallucinates symptoms based on background cues (\emph{e.g.}, inferring ``tremors'' from hospital equipment). While the \textit{Vision-Only} model improves accuracy, its rationales remain vague (\emph{e.g.}, ``walking difficulty''). In contrast, the proposed \emph{Biomechanical Tokenization Branch} successfully injects precise kinematic terminology, shifting reasoning from generic visual impressions to quantifiable observations (\emph{e.g.}, ``spastic catch'', ``asymmetric stride''). This confirms that explicit biomechanical tokenization is a prerequisite for trustworthy clinical AI.

%-------------------------------------------
\subsection{Ablation Analysis}
As detailed in the bottom section of Tab.~\ref{tab:sota}, we isolate the contribution of each modality.
Impact of Temporal Modeling: Removing the \textit{Temporal Evidence Distillation (TED)} branch causes a sharp performance drop, particularly in the \emph{Myopathic} class (F1 decreases significantly). This confirms that static VLM features struggle to capture rhythmic gait anomalies (\emph{e.g.}, waddling) without explicit temporal modeling.
Impact of Biomechanical Tokenization: Excluding the \textit{Biomechanical Branch} degrades performance on structurally defined pathologies like \emph{Parkinson's} and \emph{DCM}. The full tri-modal framework achieves the highest overall accuracy ($\textbf{\emph{68.1\%}}$), validating that textualized joint-angle tokens provide critical geometric grounding that complements visual appearance.
Interestingly, we observe that adding a single branch (\emph{e.g.}, \textit{w/o TED} or \textit{w/o Bio}) occasionally yields lower Macro F1 scores than the baseline (\textit{w/o both}). This suggests that introducing a single new modality to a frozen VLM can initially act as distractor noise. However, when both Temporal and Biomechanical cues are integrated, the model achieves a synergistic effect (Macro F1 $52.9\%$), where the structural grounding of the skeleton tokens allows the model to effectively interpret the temporal dynamics, overcoming the noise floor.

%-------------------------------------------

\subsection{Limitations and Future Work}
\label{sec:limitations}

% While BioGait-VLM demonstrates strong clinical utility, two key limitations remain. First, the Biomechanical Tokenization relies on monocular 3D pose estimation (HSMR). Although HSMR is state-of-the-art for in-the-wild video, it remains an approximation of true kinematics; precise joint angles can still be compromised by severe occlusions (e.g., loose clothing) or depth ambiguity inherent to single-view capture. Consequently, the generated tokens represent \textit{inferred} rather than \textit{measured} biomechanics.
% %
% Second, while our subject-disjoint protocol ensures rigorous evaluation, the current DCM cohort is relatively small compared to general vision datasets.
% \textbf{Future Work:} To address the kinematic fidelity gap, we have initiated Phase 2 data collection utilizing OpenCap, a multi-view markerless motion capture system. By synchronizing two calibrated smartphones, we aim to obtain research-grade musculoskeletal sequences directly, replacing the estimated SKEL parameters with verified ground truth. This will allow us to refine the tokenization logic and validate the model's sensitivity to even subtler motor deficits in a larger patient population.

%While BioGait-VLM demonstrates strong clinical utility, limitations remain. First, reliance on monocular 3D pose estimation implies that our biomechanical tokens represent \textit{inferred} rather than \textit{measured} kinematics; precise joint angles remain susceptible to single-view depth ambiguity and occlusions (\emph{e.g.}, loose clothing). Second, the current DCM cohort is relatively small.
While BioGait-VLM demonstrates clinical utility, limitations remain. First, our tokenization strictly encodes \textit{kinematic} parameters (joint angles) rather than full \textit{biomechanical} kinetics (forces/moments). Furthermore, reliance on monocular pose estimation implies these parameters are \textit{inferred} rather than measured, remaining susceptible to single-view depth ambiguity and occlusions (\emph{e.g.}, loose clothing). Second, the current DCM cohort is relatively small.
\textbf{Future Work:} We have initiated \emph{Phase 2 data collection} utilizing OpenCap~\cite{Uhlrich2023}, a multi-view markerless motion capture system. By synchronizing two calibrated smartphones, we aim to obtain research-grade musculoskeletal sequences directly, replacing estimated parameters with verified ground truth to further refine diagnostic sensitivity.
%

%--------------------------------------------------------------%
% \section{Limitations and Future Work}
% \label{sec:limitations}

%--------------------------------------------------------------%
\section{Conclusion}

We present \textbf{BioGait-VLM}, a tri-modal framework that addresses the generalization failure of standard video models by integrating explicit biomechanical reasoning into a Large Vision-Language Model (LVLM). By projecting 3D kinematics into language-aligned semantic tokens, our approach successfully decouples pathological motion from environmental shortcuts. Extensive experiments on our rigorous 8-class benchmark, which includes the newly collected DCM dataset, demonstrate state-of-the-art performance on unseen patients. Furthermore, our blinded expert study confirms that this architecture transforms the model into a transparent clinical assistant capable of generating evidence-grounded assessments. These results suggest a practical path toward interpretable, privacy-preserving digital biomarkers for neurology.

\newpage
\sloppy
\bibliographystyle{splncs04}
\bibliography{ref}

\end{document}